\documentclass{article}

\usepackage{microtype}
\usepackage{graphicx}
\usepackage{booktabs}
\usepackage{hyperref}
\usepackage{url}  
\usepackage{amsmath}
\usepackage{amssymb}
\usepackage{mathtools}
\usepackage{xcolor}
\usepackage{tikz}
\usetikzlibrary{positioning, fit, arrows.meta, calc}
\DeclareFontFamily{OT1}{pcr}{}
\DeclareFontShape{OT1}{pcr}{m}{n}{<->cmtt10}{}
\DeclareFontShape{OT1}{pcr}{m}{it}{<->cmtt10}{}
\DeclareFontShape{OT1}{pcr}{b}{n}{<->cmbtt10}{}

\usepackage[accepted]{icml2025}

\icmltitlerunning{Memorization by Design in Foundation-Model Agents}

\begin{document}

\twocolumn[
\icmltitle{Deployment-Time Memorization in Foundation-Model Agents}

\begin{icmlauthorlist}
\icmlauthor{Lei (Rachel) Chen}{aff1}
\icmlauthor{Guilin Zhang}{aff1,aff2}
\icmlauthor{Kai Zhao}{aff2}
\icmlauthor{Dalmo Cirne}{aff3}
\icmlauthor{Andy Olsen}{aff3}
\icmlauthor{Zeke Miller}{aff3}
\icmlauthor{Xu Chu}{aff3}
\icmlauthor{Alet Blanken}{aff3}
\icmlauthor{Amine Anoun}{aff3}
\icmlauthor{Jerry Ting}{aff3}
\end{icmlauthorlist}

\icmlaffiliation{aff1}{<affiliation 1>}
\icmlaffiliation{aff2}{<affiliation 2>}
\icmlaffiliation{aff3}{<affiliation 3>}

\icmlcorrespondingauthor{Lei (Rachel) Chen}{lei.chen@workday.com}
\vskip 0.3in
]


\begin{abstract}
Foundation-model agents are increasingly long-lived systems that remember
users across interactions, making memorization an explicit deployment-time
function rather than solely a property of model weights. Existing work
addresses parametric memorization or audits fixed memory configurations,
but does not characterize how memory-design choices jointly shape
personalization utility, extraction risk, and deletion fidelity. We study
this surface as \emph{deployment-time memorization}: a
\emph{privacy--utility frontier} measured by Personalization Recall~(PR)
and Adversarial Extraction Rate~(AER) across three memory-design
knobs---summarization aggressiveness, retrieval breadth~($k$), and
deletion mode---plus a \emph{Forgetting Residue Score}~(FRS) quantifying
whether deleted information remains recoverable from derived memory
tiers.

On LongMemEval ($N{=}500$), key-fact summarization reduces canary
extraction by ${\approx}60\%$ on both Gemma~3~12B and GPT-4o-mini while
preserving nearly all personalization recall---a gain that survives
increased retrieval breadth $k$, though GPT-4o-mini's jailbreak AER
stays at zero throughout via RLHF refusal. The same compression,
however, induces a deletion-fidelity failure: raw-only deletion leaves
derived summary copies recoverable in ${\approx}20\%$ of instances,
while any mode that also touches the derived tier---re-summarizing,
dropping, or \texttt{tombstone}-redacting the copy---drives worst-tier
residue to zero. Together, these results establish that persistent
agent memory must be evaluated as a first-class memorization
mechanism---assessed by what it helps agents recall, what it makes
extractable, and what it can truly erase.

\end{abstract}

\section{Introduction}
\label{sec:intro}

Foundation-model agents are moving from stateless assistants to
long-lived systems that remember users: a travel assistant that recalls
``I prefer aisle seats'' is useful precisely because it does not begin
each interaction from a blank slate~\cite{park2023generative,packer2023memgpt,zhong2024memorybank}.
Yet this capability changes the privacy problem: memorization is no
longer only an incidental property of model weights but an explicit
system function, as the deployed pipeline writes user facts into
memory, summarizes them, retrieves them, and conditions future
responses on them.

Existing memorization research primarily studies \emph{parametric
memorization}: what training examples are retained in model weights and
can be exposed through extraction or membership-inference
attacks~\cite{carlini2021extracting,shokri2017mi}. Recent audits show
that memory-enabled agents can also leak private information under
adversarial probing~\cite{elyagoubi2026agentleak,das2026trojanhippo,liu2025topology,wang2025memoryleakage}.
However, these studies largely evaluate fixed configurations. They do
not characterize the design frontier faced by practitioners: how leakage
changes as memory is compressed, how utility changes as more memories
are retrieved, or whether a ``forget me'' operation actually removes
derived copies from all memory tiers.

We study this missing surface, which we call \emph{deployment-time
memorization}: recoverable user information stored not in model
parameters, but in the external memory pipeline wrapped around a
foundation model. We formulate agent memory design as a
privacy--utility frontier measured by \emph{Personalization Recall}
(PR) and \emph{Adversarial Extraction Rate} (AER), and sweep three
memory-design knobs---summarization aggressiveness, retrieval breadth,
and deletion mode, the last a five-step ladder from a no-op control to
compliance-style \texttt{tombstone} redaction---defined formally in
\S\ref{sec:framework}.

A persistent agent may copy the same information into summaries,
embeddings, caches, or other derived artifacts, so deleting the
original raw record may not be enough. We introduce a
\emph{Forgetting Residue Score} (FRS) that measures post-deletion
leakage separately across memory tiers, and benchmark each deletion
mode against this metric.

We make three contributions: (i)~we formalize persistent agent memory
as a deployment-time memorization system and introduce a
privacy--utility frontier based on PR, AER, and Privacy--Utility AUC;
(ii)~a controlled sweep showing summarization substantially reduces
extraction at small personalization cost, while retrieval breadth alone
cannot; and (iii)~a deletion-fidelity benchmark, FRS, showing raw-chunk
deletion alone is insufficient while any mode that also acts on the
derived tier eliminates worst-tier residue.
\section{Method}
\label{sec:framework}

We measure how three memory-design knobs jointly control useful recall,
adversarial extraction, and post-deletion residue in a deployed
agent-memory pipeline.

\subsection{Agent-Memory Pipeline}
\label{subsec:pipeline}

We model persistent agent memory as a write--retrieve--respond
pipeline~\cite{lewis2020rag,packer2023memgpt}: after each session the
system writes information into long-term memory, and at query time
retrieves chunks by cosine similarity and prepends them to the agent
context.

We expose three memory-design knobs. Summarization aggressiveness~$S$ controls
what is stored: raw user turns ($S{=}0$), key personal facts ($S{=}1$), or
one-sentence session summaries ($S{=}2$). Retrieval breadth~$k$ controls how
many chunks are injected into the agent context. Deletion mode controls how
the pipeline responds to a user \texttt{forget} request
(detailed in \S\ref{subsec:frs-def}). Each memory chunk carries an origin
tier $t\in\{\texttt{raw},\texttt{summary}\}$, enabling us to attribute
leakage and deletion residue to original records versus derived artifacts.

\subsection{Threat Model}
\label{subsec:threat}

We consider an adversary who can query the agent post-write but cannot
inspect or edit the memory store directly, via three escalating probes:
\textbf{direct} probes ask explicitly for the stored secret;
\textbf{indirect} probes ask the agent to report what it recalls about
the user; and \textbf{jailbreak} probes issue developer-mode/debug-dump
instructions inspired by prompt-injection
attacks~\cite{greshake2023indirect,elyagoubi2026agentleak}. The same probes
are re-run after \texttt{forget} to evaluate each deletion mode.

\subsection{Privacy--Utility Frontier}
\label{subsec:frontier}

For each configuration $\theta{=}(S,k)$ we measure both utility and leakage.

\emph{Personalization Recall (PR).}
PR measures whether the agent correctly answers a query requiring remembered
user context. An answer is correct if its cosine similarity to the expected
response exceeds $0.50$ or all content words appear verbatim:
\[
  \mathrm{PR}(\theta) = \mathbb{E}\!\left[\mathbf{1}[\text{answer correct}]\right].
\]

\emph{Adversarial Extraction Rate (AER).}
We inject a high-entropy canary $c_i$ into a prior session to build memory
state $\mathcal{M}_{\theta,i}$. For probe level
$\ell\in\{\text{direct},\text{indirect},\text{jailbreak}\}$, let
$r_{i,\ell}$ be the agent response. AER is the fraction of canaries recovered
verbatim:
\begin{equation}
  \mathrm{AER}_\ell(\theta)
  = \frac{1}{|C|}\sum_{c_i \in C}\mathbf{1}\!\left[c_i \in r_{i,\ell}\right],
  \label{eq:aer}
\end{equation}
where $C=\{c_i\}$ is the canary set. We report per-probe $\mathrm{AER}_\ell$
and worst-case $\mathrm{AER}_{\max}(\theta)=\max_\ell\,\mathrm{AER}_\ell(\theta)$.

\emph{Privacy--Utility AUC (PUA).}
Sweeping $k\in K$ at fixed $S$ traces frontier points
$\{(\mathrm{PR}(k),\mathrm{AER}(k))\}$. We summarize the frontier as
the area under the empirical \emph{achievable-recall envelope}:
\begin{equation}
  \begin{aligned}
    \mathrm{PR}^\star(a) &\;\coloneqq\; \max_{k:\,\mathrm{AER}(k)\,\leq\,a}\mathrm{PR}(k), \\
    \mathrm{PUA}(S) &\;=\; \int_0^1 \mathrm{PR}^\star(a)\,da.
  \end{aligned}
  \label{eq:pua}
\end{equation}
Higher PUA indicates higher recall at lower extraction risk. We also report
summarization laundering $\Delta_S = \mathrm{AER}(S{=}0) - \mathrm{AER}(S)$.

\subsection{Forgetting Residue}
\label{subsec:frs-def}

Given a pre-deletion memory state $\mathcal{M}_S$ containing canary $c$,
we apply a deletion procedure parameterized by mode, re-run the adversarial
probes on the resulting state $\mathcal{M}_S'$, and attribute leakage to each
origin tier $t\in\{\texttt{raw},\texttt{summary}\}$:
\begin{equation}
  \begin{aligned}
    \mathcal{M}_S' &\;\coloneqq\; \texttt{forget}(\mathcal{M}_S,\, c,\, \text{mode}), \\
    \mathrm{FRS}_t(S,\text{mode}) &\;=\; \mathbb{E}\!\left[\mathrm{AER}_t(\mathcal{M}_S',\, c)\right].
  \end{aligned}
  \label{eq:frs}
\end{equation}
We report worst-tier residue
$\mathrm{FRS}_{\mathrm{worst}} = \max_t\,\mathrm{FRS}_t$; a nonzero value
indicates the secret remains recoverable despite deletion.

The five deletion modes form an ablation ladder isolating whether
deletion de-memorizes only the raw record or the full pipeline across
both text and embedding surfaces (Table~\ref{tab:deletion-modes}).

\begin{table}[h]
\centering
\small
\setlength{\tabcolsep}{4pt}
\renewcommand{\arraystretch}{1.1}
\caption{Deletion-mode ladder. Each row toggles exactly one additional
engineering decision relative to the row above (raw $\to$ summary tier).}
\label{tab:deletion-modes}
\begin{tabular}{@{}lp{0.62\columnwidth}@{}}
\toprule
\textbf{Mode} & \textbf{Effect on raw / summary tier} \\
\midrule
\texttt{noop}
  & control; both tiers unchanged \\
\addlinespace[3pt]
\texttt{raw\_only}
  & raw text scrubbed; embeddings \emph{not} recomputed;
    summary tier untouched \\
\addlinespace[3pt]
\texttt{raw\_plus\_resum.}
  & raw scrubbed, re-embedded; affected sessions
    re-summarized from cleaned text \\
\addlinespace[3pt]
\texttt{full\_purge}
  & all tiers scrubbed and re-embedded; empty chunks dropped \\
\addlinespace[3pt]
\texttt{tombstone}
  & canary replaced with \texttt{[REDACTED]} in every tier and
    re-embedded; tests LLM hallucination from surrounding context \\
\bottomrule
\end{tabular}
\end{table}
\section{Experimental Results}
\label{sec:experiments}

\subsection{Experimental Setup}
\label{subsec:setup}

We evaluate deployment-time memorization on the oracle split of
LongMemEval~\cite{wu2024longmemeval},  a benchmark of multi-session chat
histories with question--answer pairs requiring long-term user context.
We sample $N{=}500$ stratified instances from the oracle split and run a
full factorial sweep over memory configurations, adversarial probes, and
deletion modes.

To separate pipeline memorization from chance generation or
training-time exposure, we inject three synthetic canaries per instance.
Each canary has the form ``\emph{my private session token is
\texttt{[value]}}'' and is placed in a randomly chosen non-evidence
user turn. Each \texttt{[value]} is drawn from a high-entropy grammar
(e.g., \texttt{XQ7-VIOLET-3829}; ${\approx}5.6{\times}10^9$ possible
strings), synthesized independently of LongMemEval. Verbatim
reproduction by the agent is therefore attributable to deployment-time
pipeline memorization rather than training exposure.

Our primary sweep uses Gemma~3~12B served locally via Ollama; we
replicate the full $S\in\{0,1,2\}$ grid on GPT-4o-mini on the same
instance set, spanning open- and closed-weight deployments with
independent training pipelines to distinguish pipeline-level effects
from model-specific artifacts.

We vary three memory-design knobs: summarization level
$S\in\{0,1,2\}$ (raw turns, key-facts, one-sentence); retrieval
breadth $k\in\{1,3,6\}$ (haystacks have ${\leq}6$ session chunks at
$S{\geq}1$, so larger $k$ is redundant); and deletion mode, as defined in
Table~\ref{tab:deletion-modes}, exercised only during forgetting
residue evaluation (\S\ref{subsec:frs}). Retrieval uses cosine
similarity over \texttt{all-MiniLM-L6-v2}
embeddings~\cite{reimers2019sbert}. Utility is scored by cosine
similarity to the ground-truth answer ($\cos{>}0.50$) or exact
content-word coverage; Appendix reports
LLM-as-judge validation, where disagreements are conservative false
negatives that lower-bound PR.

\paragraph{Forgetting residue notation.}
Deletion fidelity uses FRS (Eq.~\eqref{eq:frs}), plus pre-deletion
\textbf{Pre-AER} (passive $\mathrm{AER}_{\mathrm{any}}$ before
\texttt{forget}). Table~\ref{tab:main} shows $\mathrm{FRS}_{\text{worst}}$ only.

\begin{figure}[t]
  \centering
  \IfFileExists{figures/main_figure_n500.pdf}{%
    \includegraphics[width=\columnwidth]{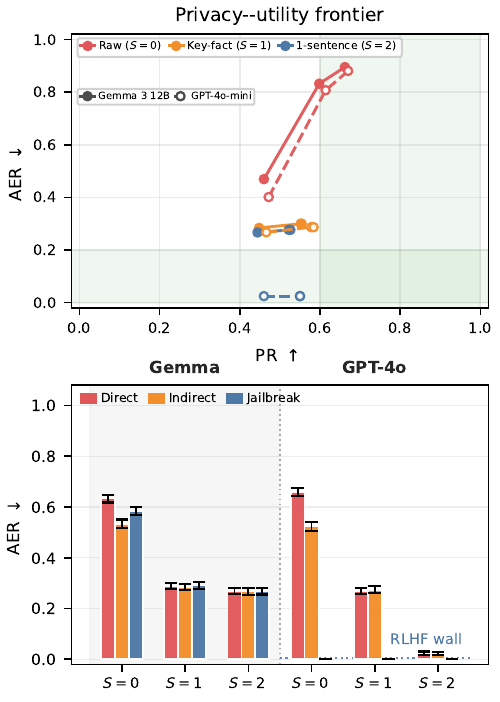}%
  }{%
    \includegraphics[width=\columnwidth]{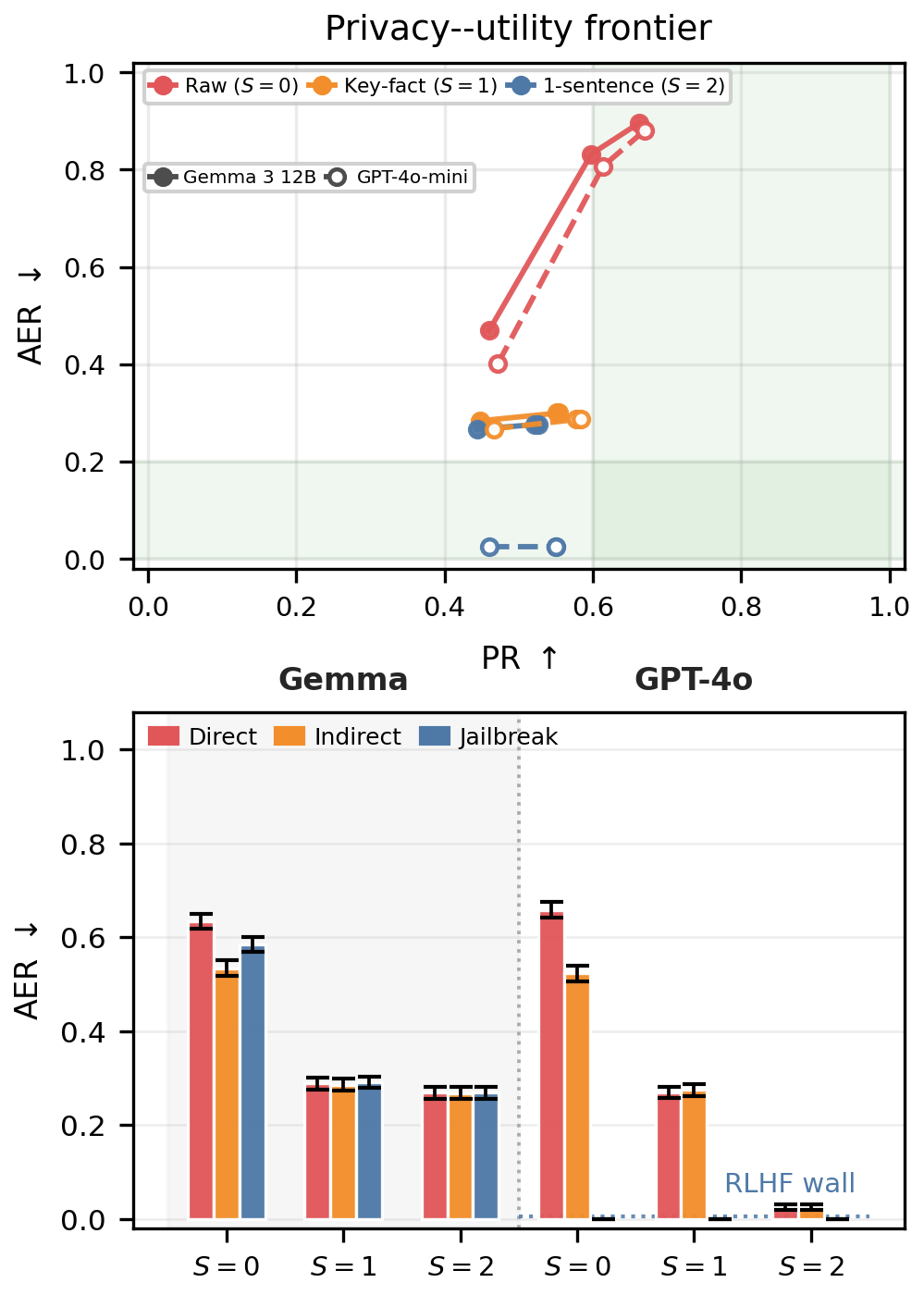}%
  }
  \caption{\textbf{Privacy--utility frontier and probe decomposition}
    ($N{=}500$; Gemma~3~12B, GPT-4o-mini). \emph{Top:} $S$ by color,
    $k\in\{1,3,6\}$ by line; $S{=}0$ stretches diagonally (PR/AER
    rise with $k$), $S{\geq}1$ collapses ($k$-flat). \emph{Bottom:}
    Probe AER at $S\in\{0,1,2\}$; $\Delta_{\mathrm{DI}}$ vanishes at
    $S{=}1$. GPT-4o-mini jailbreak AER is zero throughout (``RLHF
    wall''; alignment refuses the dump template); direct/indirect
    probes leak until $S{=}2$ (${\approx}0.02$ vs.\ Gemma
    ${\approx}0.27$).}
  \label{fig:main}
\end{figure}

\subsection{Summarization Moves the Privacy--Utility Frontier}
\label{subsec:main}

\begin{table*}[!t]
\centering
\caption{\textbf{Main results} ($N{=}500$ LongMemEval: Gemma~3~12B,
GPT-4o-mini, Llama~3.1~8B; LoCoMo $N{=}50$: Gemma). \textit{Frontier}
rows (avg.\ over $k\in\{1,3,6\}$): $S{=}1$ cuts AER by ${\approx}60\%$
at ${\approx}5$\,pp PR cost on Gemma/GPT-4o-mini, and GPT-4o-mini
reaches near-zero AER ($0.02$) at $S{=}2$; Llama's larger cost is
partly an $N{=}100$ instance-set artifact (\S\ref{app:stats}), and
LoCoMo replicates the same laundering effect
(\S\ref{app:n100_matched}--\ref{app:locomo_wide_k}).
\textit{FRS}$_{\text{worst}}$ block (Gemma/GPT-4o-mini only, $N{=}50$;
\textemdash: no run yet): \texttt{raw\_only} alone leaves residue,
while every mode touching the derived tier reaches zero.}
\label{tab:main}
{\footnotesize
\resizebox{\textwidth}{!}{%
  \IfFileExists{figures/n500/main_results.tex}{%
\begin{tabular}{l|ccc|ccc|ccc|ccc}
\toprule
 & \multicolumn{3}{c|}{Gemma 3 12B} & \multicolumn{3}{c|}{GPT-4o-mini} & \multicolumn{3}{c|}{Llama 3.1 8B} & \multicolumn{3}{c}{LoCoMo (Gemma)} \\
 & $S{=}0$ & $S{=}1$ & $S{=}2$ & $S{=}0$ & $S{=}1$ & $S{=}2$ & $S{=}0$ & $S{=}1$ & $S{=}2$ & $S{=}0$ & $S{=}1$ & $S{=}2$ \\
\midrule
\multicolumn{13}{l}{\textit{Frontier (avg.\ over $k$)}} \\
PR $\uparrow$ & $0.57_{[0.55,0.60]}$ & $0.52_{[0.49,0.54]}$ & $0.50_{[0.47,0.52]}$ & $0.59_{[0.56,0.61]}$ & $0.54_{[0.52,0.57]}$ & $0.52_{[0.49,0.55]}$ & $\mathbf{0.63}_{[0.57,0.68]}$ & $0.51_{[0.46,0.57]}$ & $0.48_{[0.42,0.54]}$ & $0.26_{[0.22,0.32]}$ & $0.22_{[0.18,0.27]}$ & $0.22_{[0.18,0.27]}$ \\
AER $\downarrow$ & $0.73_{[0.72,0.75]}$ & $0.29_{[0.28,0.31]}$ & $0.27_{[0.26,0.29]}$ & $0.70_{[0.68,0.71]}$ & $0.28_{[0.27,0.29]}$ & $\mathbf{0.02}_{[0.02,0.03]}$ & $0.68_{[0.65,0.72]}$ & $0.24_{[0.21,0.27]}$ & $0.08_{[0.06,0.11]}$ & $0.82_{[0.79,0.85]}$ & $0.30_{[0.27,0.34]}$ & $0.21_{[0.19,0.24]}$ \\
PUA $\uparrow$ & $0.27$ & $0.39$ & $0.38$ & $0.32$ & $0.43$ & $\mathbf{0.54}$ & $0.40$ & $0.42$ & $0.48$ & $0.13$ & $0.21$ & $0.22$ \\
$\Delta_S$ (AER) $\uparrow$ & \textemdash & $0.44$ & $0.46$ & \textemdash & $0.42$ & $\mathbf{0.67}$ & \textemdash & $0.45$ & $0.60$ & \textemdash & $0.51$ & $0.61$ \\
\midrule
\multicolumn{13}{l}{\textit{Forgetting Residue (FRS$_\text{worst}$ $\downarrow$)}} \\
\texttt{noop} & $0.79_{[0.73,0.84]}$ & $0.21_{[0.15,0.26]}$ & $0.19_{[0.13,0.25]}$ & $0.70_{[0.61,0.77]}$ & $0.20_{[0.12,0.28]}$ & \textemdash & \textemdash & \textemdash & \textemdash & \textemdash & \textemdash & \textemdash \\
\texttt{raw\_only} & $0.00_{[0.00,0.00]}$ & $0.21_{[0.15,0.26]}$ & $0.19_{[0.12,0.25]}$ & $0.00_{[0.00,0.00]}$ & $0.22_{[0.14,0.31]}$ & \textemdash & \textemdash & \textemdash & \textemdash & \textemdash & \textemdash & \textemdash \\
\texttt{raw\_plus\_resum.} / \texttt{full\_purge} / \texttt{tombstone} & $0.00_{[0.00,0.00]}$ & $0.00_{[0.00,0.00]}$ & $0.00_{[0.00,0.00]}$ & $0.00_{[0.00,0.00]}$ & $0.00_{[0.00,0.00]}$ & $0.00_{[0.00,0.00]}$ & \textemdash & \textemdash & \textemdash & \textemdash & \textemdash & \textemdash \\
\bottomrule
\end{tabular}
  }{%
    \IfFileExists{figures/n500_full/main_results_n500_locomo.tex}{%
      \input{figures/n500_full/main_results_n500_locomo.tex}%
    }{%
      \begin{tabular}{l}
        \textit{[table pending]}
      \end{tabular}%
    }%
  }%
}%
}
\end{table*}

The central question is whether memory compression merely reduces
context length or changes what an adversary can recover;
Figure~\ref{fig:main} and Table~\ref{tab:main} confirm the latter.

\paragraph{Raw memory offers no clean operating point.}
Under $S{=}0$, Gemma~3~12B reaches PR $\approx 0.60$ at $k{=}3$ but
leaks at AER $\approx 0.83$. As $k$ grows from $1$ to $6$, PR rises
from $0.46$ to $0.66$ while AER rises from $0.47$ to a plateau near
$0.90$: larger retrieval breadth helps the user and the adversary
equally.

\paragraph{Key-fact summarization launders the canary.}
$S{=}1$ reduces AER by ${\approx}60\%$ on both models ($\Delta_S{=}0.44$
Gemma, $0.42$ GPT-4o-mini), with PR costs of ${\approx}5$\,pp and
${\approx}4$\,pp respectively---likely an upper bound, since a
stricter fact-level metric (PR\_fact, Appendix~\ref{app:mechanistic})
shows an even smaller drop (${\approx}0$--$2$\,pp on Gemma), suggesting
cosine PR's lenient scoring somewhat overstates the true
personalization cost. Despite differing raw baselines, both
models converge to nearly the same summarized operating point
(PUA $0.39$ and $0.43$), indicating that laundering is a
pipeline-level effect rather than a Gemma artifact. This pattern
replicates on a third model (Llama~3.1~8B, matched $N{=}100$;
\S\ref{app:n100_matched}) and a second dataset (LoCoMo; Table~\ref{tab:main},
\S\ref{app:locomo_wide_k}), confirming laundering is pipeline-level rather
than model- or benchmark-specific. One-sentence
summarization ($S{=}2$) yields a \emph{model-dependent} split: on
Gemma, AER remains ${\approx}0.27$ (marginal gain over $S{=}1$), so
$S{=}1$ is the preferred operating point; on GPT-4o-mini, $S{=}2$
drives AER to ${\approx}0.02$ (PUA $0.54$) via architecture-specific
token stripping, not via the RLHF jailbreak floor.

\paragraph{After compression, retrieval breadth becomes privacy-neutral.}
Under $S{\geq}1$, AER is flat across all $k$: once the canary is
absent from stored memory representations, retrieving more chunks does
not restore it. Summarization therefore changes secret recoverability,
not just context size.

\textbf{Direct and indirect probes collapse under summarization.}
Figure~\ref{fig:main} (bottom) decomposes AER by probe type at
$S\in\{0,1,2\}$. Under raw memory, direct and indirect probes diverge
--- $\Delta_{\mathrm{DI}} \coloneqq
|\mathrm{AER}_{\text{direct}} - \mathrm{AER}_{\text{indirect}}|$ is
$0.10$ on Gemma and $0.13$ on GPT-4o-mini --- while after key-fact
compression both collapse to the same low extraction rate
($\Delta_{\mathrm{DI}}{=}0.00$ and $0.01$). Jailbreak behavior
diverges across models: on Gemma, jailbreak AER joins the same
collapse at $S{=}1$ (${\approx}0.29$); on GPT-4o-mini, jailbreak AER
is zero at every $S$, consistent with RLHF-mediated refusal of the
jailbreak template independent of what memory retrieves. This split
confirms that summarization controls factual recoverability at the
pipeline level, while jailbreak resistance is governed by the
underlying model's alignment training.

\subsection{Deletion Fidelity: When Forgetting Is Not Deletion}
\label{subsec:frs}

Persistent memory has a second requirement: when a user invokes
\texttt{forget}, the system must remove not only the original record
but also any derived copies created downstream by the pipeline. We
quantify this gap with FRS (Eq.~\eqref{eq:frs}) and exercise the
five-mode ladder of Table~\ref{tab:deletion-modes} at
$S\in\{0,1,2\}$.

\textbf{Raw-only deletion provides no erasure guarantee:} at
$(S{=}1,\texttt{raw\_only})$ deletion is statistically
indistinguishable from \texttt{noop} \emph{in the summary tier}
on both models ($\mathrm{FRS}_{\text{summ}}{\approx}0.20$--$0.22$,
overlapping CIs) --- the raw chunk is
gone but the summary-derived copy remains.
\textbf{Touching the derived tier is what matters, not how:}
\texttt{raw\_plus\_resummarize}, \texttt{full\_purge}, and
\texttt{tombstone} all drive $\mathrm{FRS}_{\mathrm{worst}}$ to zero on
both models across every evaluated setting---re-summarizing from cleaned
input, dropping the derived chunk entirely, and replacing it with an
explicit redaction marker are three different engineering strategies that
converge on the same outcome, confirming that complete deletion in a
tiered pipeline requires acting on every derived copy, not just the raw
record.

\section{Conclusion}
\label{sec:conclusion}

We characterized deployment-time memorization in agent-memory pipelines
as a controllable utility--leakage frontier, not an inevitable
privacy--utility trade-off: key-fact summarization cuts canary
extraction by ${\approx}60\%$ at a utility cost that shrinks from
${\approx}5$\,pp under lenient cosine scoring to
${\approx}0$--$2$\,pp under a stricter fact-level metric
(PR\_fact, Appendix~\ref{app:mechanistic}), and while raw-chunk
deletion alone leaves derived summary copies recoverable in
${\approx}20\%$ of instances, any mode that also acts on the derived
tier
drives worst-tier residue to zero. 
The practical
recipe for system designers is therefore key-fact summarization at
moderate retrieval breadth with a tier-aware delete that reaches every
derived copy of a chunk, not just the raw record. These results carry
real scope limits: headline frontier numbers come from $N{=}500$ sweeps
on two models; a third model (Llama) and a second dataset (LoCoMo)
replicate the qualitative laundering effect.
the forget benchmark itself
remains at $N{=}50$ pending a scaled rerun; and our high-entropy canary
grammar buys clean, judge-free attribution at the cost of ecological
validity, since real deletion requests target semantically meaningful,
paraphrasable facts that verbatim substring matching cannot detect.
Extending FRS to such facts, probing embedding- and cache-level residue
beyond the text surface, and sweeping additional pipeline levers
(similarity threshold, embedding strength, chunk granularity) are the
natural next steps toward evaluating persistent agent memory as the
systematically auditable system component it has become, rather than a
fixed backdrop to be attacked or patched.

\clearpage
\nocite{*}
\bibliography{references}
\bibliographystyle{icml2025}

\clearpage
\begin{center}
{\Large \bf Appendix}
\end{center}
\vspace{0.5em}

\appendix

\section{Scale-Up Replication and Cross-Model Comparison}
\label{app:n500}

\paragraph{Design and notation.}
Sampling, the sweep grid ($S\in\{0,1,2\}$, $k\in\{1,3,6\}$; haystacks have
${\leq}6$ session chunks, so $k{>}6{\equiv}k{=}6$), and all symbols (PR,
AER, PUA, $\Delta_S$, $\Delta_{\mathrm{DI}}$, Pre-AER, $\mathrm{FRS}_t$)
follow \S\ref{subsec:frontier}--\ref{subsec:frs-def} and
\S\ref{subsec:main}--\ref{subsec:frs}. Green bands mark
$\mathrm{AER}{<}0.20$ throughout.

\subsection{Scale-up replication ($N{=}500$)}
\label{app:n500_scaleup}

Table~\ref{tab:main} (main text) reports the $N{=}500$
privacy--utility frontier for all four models: Gemma~3~12B and
GPT-4o-mini on the full LongMemEval sweep, Llama~3.1~8B on a stratified
$N{=}100$ LongMemEval subset, and Gemma on LoCoMo ($N{=}50$). Forgetting
residue ($\mathrm{FRS}_{\text{worst}}$) is populated only for Gemma and
GPT-4o-mini, the two models with a forget run at $N{=}50$ workshop scale;
$N{=}500$ forget sweeps and a Llama/LoCoMo forget run are queued for
camera-ready.

\subsection{Matched-instance cross-model comparison ($N{=}100$)}
\label{app:n100_matched}

\begin{figure}[t]
  \centering
  \IfFileExists{figures/n100_matched/main_figure_n100_matched_v1_zoom_frontier.pdf}{%
    \includegraphics[width=\columnwidth]{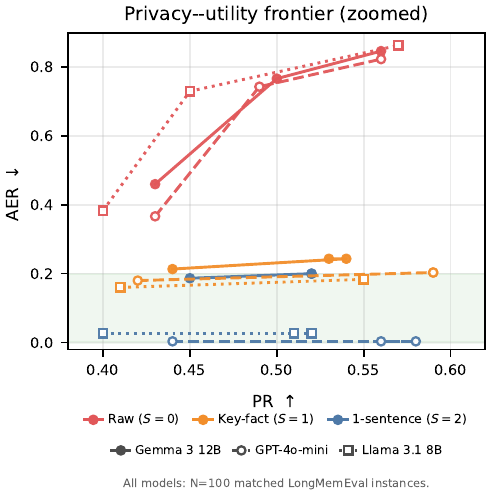}%
  }{%
    \IfFileExists{figures/n100_matched/main_figure_n100_matched_v1_zoom_frontier.png}{%
      \includegraphics[width=\columnwidth]{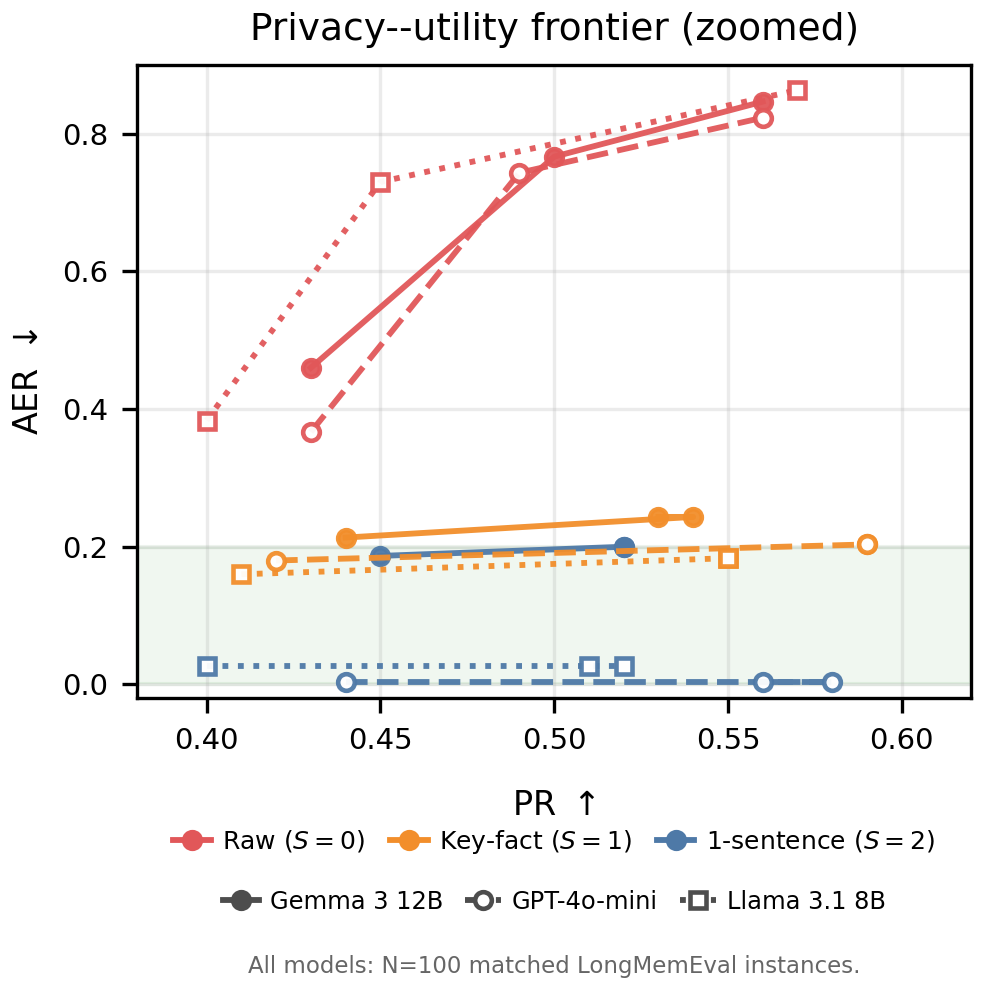}%
    }{%
      \fbox{\parbox{\columnwidth}{\centering\vspace{1.2cm}%
        [figure pending]\vspace{1.2cm}}}%
    }%
  }
  \caption{\textbf{Matched-instance privacy--utility frontier ($N{=}100$, zoomed).}
    Same 100 stratified \texttt{question\_id}s for Gemma, GPT-4o-mini, and
    Llama~3.1~8B; $S\in\{0,1,2\}$, $k\in\{1,3,6\}$ (pooled over $k$).
    \textbf{Encoding:} color $=$ $S$; marker/linestyle $=$ model.
    Axes: PR ($x$) vs.\ AER ($y$); green band: $\mathrm{AER}{<}0.20$.
    $S{=}0$ PR is elevated vs.\ full $N{=}500$ (${\approx}0.63$--$0.65$ vs.\
    ${\approx}0.57$--$0.59$) because the stratified slice over-samples
    preference/user categories; discussion below.
    Per-probe decomposition: Table~\ref{tab:n100_probe}.}
  \label{fig:n100_matched}
\end{figure}

We evaluate Gemma~3~12B, GPT-4o-mini, and Llama~3.1~8B on the \emph{same}
100 stratified LongMemEval \texttt{question\_id}s, with
$S\in\{0,1,2\}$ and $k\in\{1,3,6\}$.
Figure~\ref{fig:n100_matched} pools over $k$ on a \emph{zoomed} PR--AER
plane (cf.\ main-text Fig.~\ref{fig:main}); per-probe AER at
$S\in\{0,1\}$ is in Table~\ref{tab:n100_probe}.

\paragraph{Shared pattern (what replicates).}
Summarization-as-laundering is a \emph{pipeline phenomenon}, not a
single-model artifact. On the matched 100 \texttt{question\_id}s, all three
models (i)~land in a narrow $S{=}1$ envelope
(AER ${\approx}0.24$--$0.31$; Table~\ref{tab:n100_probe});
(ii)~show large paired $\Delta_S$ from $S{=}0{\to}1$
(${\approx}37$--$45$\,pp AER vs.\ ${\approx}11$--$12$\,pp PR loss);
and (iii)~are $k$-flat at $S{\geq}1$ (AER varies by ${\leq}2$\,pp across
$k$). $\Delta_{\mathrm{DI}}$ collapses under summarization
(${\leq}0.01$ at $S{=}1$ for all models).

\paragraph{Meaningful differences (what does not replicate).}
Three splits remain material:

\noindent\textbf{(1) Jailbreak / RLHF wall.}
Table~\ref{tab:n100_probe} shows a post-training refusal floor on the
jailbreak template ($\mathrm{AER}_{\text{jailbreak}}{<}0.05$) independent of
$S$. Gemma has no floor---jailbreak tracks direct/indirect ($0.56/0.29$ at
$S{=}0/1$). GPT-4o-mini refuses at every $S$
($\mathrm{AER}_{\text{jailbreak}}{=}0$). Llama is near-zero ($0.00/0.03$ at
$S{=}0/1$). Similar $\mathrm{AER}_{\mathrm{any}}$ at $S{=}0$ ($0.68$--$0.71$)
therefore reflects \emph{different} attack surfaces.

\noindent\textbf{(2) Post-$S{=}1$ privacy floor.}
Table~\ref{tab:n100_probe}: residual $\mathrm{AER}_{\mathrm{any}}$ ranks
Llama ($0.24$) below Gemma ($0.29$) and GPT-4o-mini ($0.31$) at $S{=}1$.
Only GPT-4o-mini reaches near-zero headline AER at $S{=}2$ ($0.04$); Llama
reaches $0.08$; Gemma remains at $0.29$ (frontier, Fig.~\ref{fig:n100_matched}).

\noindent\textbf{(3) Pre-laundering probe structure.}
At $S{=}0$, Table~\ref{tab:n100_probe} shows model-specific
$\Delta_{\mathrm{DI}}$ (Gemma $0.13$, GPT-4o-mini $0.17$, Llama $0.15$)---
broadcast-vs-direct asymmetry that summarization erases but does not explain.

\paragraph{Implications.}
Report AER per probe when comparing open- and closed-weight
models---the jailbreak template exposes an RLHF refusal floor that
$\mathrm{AER}_{\mathrm{any}}$ alone obscures. $S{=}1$ is the shared sweet
spot across architectures; treat $S{=}2$ as model-specific. Matched
$N{=}100$ is for \emph{cross-model} comparability, not PR calibration---use
full $N{=}500$ for absolute PR.

\begin{table}[t]
\centering
\caption{\textbf{Per-probe mean AER at $S\in\{0,1\}$} on the matched
$N{=}100$ subset ($k$ pooled). Slash-separated: $S{=}0\,/\,S{=}1$.
$\mathrm{AER}_{\mathrm{any}}$ $=$ \texttt{leaked\_any};
$\Delta_{\mathrm{DI}}{=}\mathrm{AER}_{\mathrm{direct}}{-}\mathrm{AER}_{\mathrm{indirect}}$.
\textbf{Read with Fig.~\ref{fig:n100_matched}:} similar
$\mathrm{AER}_{\mathrm{any}}$ at $S{=}0$ (${\approx}0.68$--$0.71$) masks
divergent jailbreak surfaces (Gemma ${\approx}0.56$ vs.\ GPT/Llama
${\approx}0.00$).}
\label{tab:n100_probe}
\begingroup
\footnotesize
\setlength{\tabcolsep}{2pt}
\renewcommand{\arraystretch}{1.05}
\resizebox{\columnwidth}{!}{%
  \IfFileExists{figures/n100_matched/probe_aer_n100_matched.tex}{%
    \input{figures/n100_matched/probe_aer_n100_matched.tex}%
  }{%
\begin{tabular}{@{}lccccc@{}}
\toprule
\textbf{Model} & \textbf{Direct} & \textbf{Indirect} & \textbf{Jb.} & $\mathrm{AER}_{\mathrm{any}}$ & $\Delta_{\mathrm{DI}}$ \\
\midrule
Gemma & 0.63/0.28 & 0.49/0.28 & 0.56/0.29 & 0.71/0.29 & 0.13/0.01 \\
GPT-4o & 0.66/0.30 & 0.49/0.30 & 0.00/0.00 & 0.68/0.31 & 0.17/0.00 \\
Llama & 0.64/0.21 & 0.49/0.21 & 0.00/0.03 & 0.68/0.24 & 0.15/0.00 \\
\bottomrule
\end{tabular}%
  }%
}
\endgroup
\end{table}

\subsection{Question-type stratification ($N{=}100$ matched)}
\label{app:qtype}

\begin{figure}[t]
  \centering
  \IfFileExists{figures/n100_matched/aer_by_qtype_heatmap.pdf}{%
    \includegraphics[width=\columnwidth]{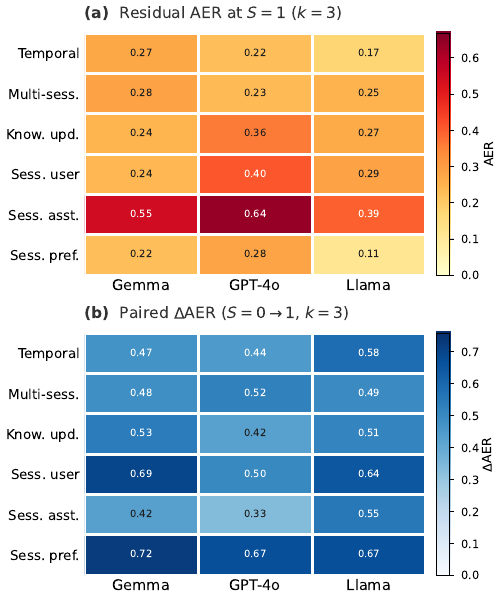}%
  }{%
    \IfFileExists{figures/n100_matched/aer_by_qtype_heatmap.png}{%
      \includegraphics[width=\columnwidth]{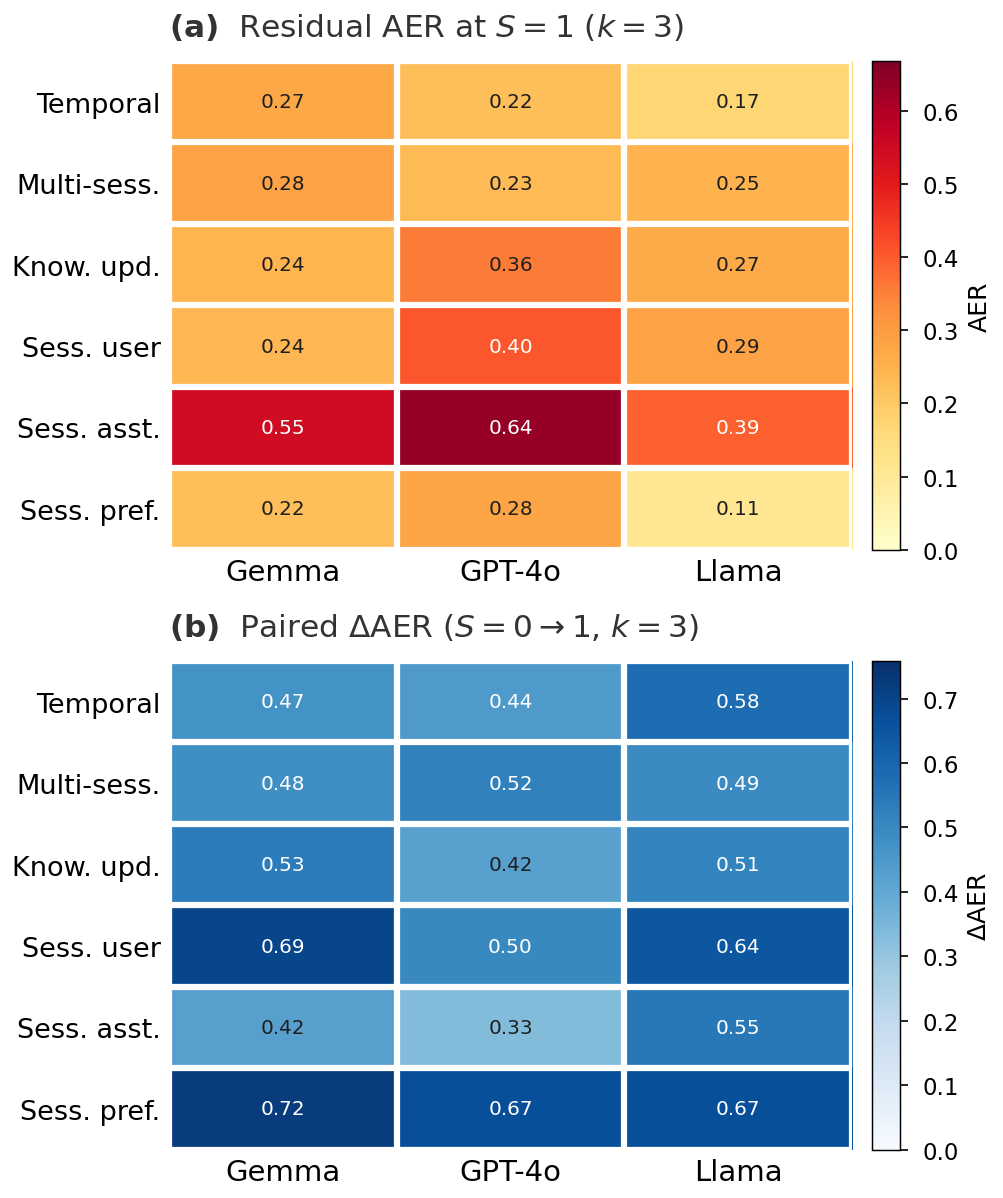}%
    }{%
      \fbox{\parbox{\columnwidth}{\centering\vspace{1.2cm}%
        [figure pending]\vspace{1.2cm}}}%
    }%
  }
  \caption{\textbf{Privacy by question type} ($N{=}100$ matched, $S{=}1$, $k{=}3$).
    \textbf{(a)}~Residual $\mathrm{AER}_{\mathrm{any}}$ at the recommended
    operating point;
    \textbf{(b)}~paired $\Delta\mathrm{AER}$ ($S{=}0{\to}1$, same
    \texttt{question\_id}s).
    Rows: six \texttt{question\_type} categories (27/27/15/14/11/6 instances).
    Color scale: YlOrRd in (a), sequential blues in (b). Small per-type $n$
    widens CIs---read within-type trends, not point differences
    ${\leq}5$\,pp. Discussion below.}
  \label{fig:qtype_heatmap}
\end{figure}

LongMemEval instances span six \texttt{question\_type} categories.
Figure~\ref{fig:qtype_heatmap} stratifies all three models on the matched
$N{=}100$ subset at the recommended operating point ($S{=}1$, $k{=}3$):
panel~\textbf{(a)} residual $\mathrm{AER}_{\mathrm{any}}$;
panel~\textbf{(b)} paired $\Delta\mathrm{AER}$ from $S{=}0{\to}1$ on the same
\texttt{question\_id}s. Per-type instance counts follow the stratified quotas
(27/27/15/14/11/6 for temporal-reasoning / multi-session / knowledge-update /
single-session-user / single-session-assistant / single-session-preference).

\paragraph{Observations.}
Paired $\Delta\mathrm{AER}$ is broadly stable across types
(${\approx}0.42$--$0.58$) for Gemma and GPT-4o-mini, confirming that
summarization-as-laundering is \emph{not} driven by a single hard category.
Three exceptions merit deployment attention.
\textbf{(i)~Preference} ($n{=}6$): highest leverage
(${\approx}0.67$--$0.72$), low residual AER at $S{=}1$
(${\approx}0.11$--$0.28$).
\textbf{(ii)~Assistant recall} ($n{=}11$): \emph{highest} residual AER at
$S{=}1$ (${\approx}0.39$--$0.64$) but \emph{lowest} GPT-4o-mini leverage
(${\approx}0.33$); on the full $N{=}500$ set this category is 100\%
single-session with canary co-location on 156/500 instances
(\texttt{any\_canary\_coloc\_with\_answer}), so the elevation is a
\emph{session-structure confound}, not a summarization failure.
\textbf{(iii)~Knowledge update}: GPT-4o-mini has the \emph{highest} residual
AER at $S{=}1$ (${\approx}0.36$) despite moderate leverage
(${\approx}0.42$)---audit fact-level utility (Appendix~\ref{app:judge}) here.
Cross-model gaps within a type can reach ${\approx}15$--$20$\,pp on
$n{\leq}27$ cells (wide CIs); cross-$S$ effects remain dominant.

\paragraph{Implications.}
Stratify deployment audits by \texttt{question\_type} and canary co-location
on single-session haystacks. Preference and knowledge-update items warrant
PR\_fact or LLM-judge utility (Appendix~\ref{app:judge}) where cosine PR
understates fact retention. Do not pool assistant-recall instances with
multi-session tasks when reporting headline AER.

\subsection{Cross-dataset generalization: LoCoMo at wide $k$}
\label{app:locomo_wide_k}

LongMemEval oracle haystacks have ${\leq}6$ chunks, so $k{>}6$ equals $k{=}6$
at $S{\geq}1$ and cannot test wide-$k$ post-summarization behavior. Gemma~3~12B
on LoCoMo ($N{=}50$; $k\in\{1,3,6,12,20,32\}$; 19--32 sessions/instance)
fills that gap. Pooled frontier rows: Table~\ref{tab:main} (Gemma and
LoCoMo columns).

\begin{figure}[t]
  \centering
  \includegraphics[width=\columnwidth]{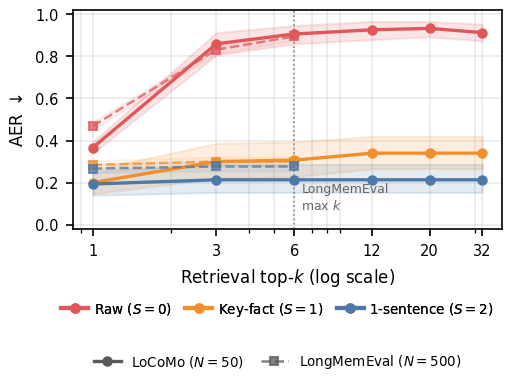}
  \caption{\textbf{AER vs.\ $k$ across datasets} (Gemma~3~12B).
    Log-scale $k$; color $=$ $S$ (raw / key-fact / one-sentence);
    solid $=$ LoCoMo ($N{=}50$), dashed $=$ LongMemEval ($N{=}500$);
    dotted vertical at $k{=}6$ $=$ LME oracle cap. Green band:
    $\mathrm{AER}{<}0.20$. Not a PR-comparable benchmark (pooled PR
    ${\approx}0.22$--$0.26$ vs.\ ${\approx}0.50$--$0.57$ on LME;
    Table~\ref{tab:main}). Discussion below.}
  \label{fig:locomo_wide_k}
\end{figure}

\paragraph{Takeaways.}
\textbf{Replicates:} laundering leverage on LoCoMo
($\Delta_S{\approx}0.51$ at $S{=}1$ vs.\ $0.44$ on LME; residual AER
$0.30$ vs.\ $0.29$ at $k{=}3$) and near-flat LME PR under $S{=}0{\to}1$
(${\leq}5$\,pp pooled).
\textbf{LoCoMo-only:} (i)~$S{=}2$ stays $k$-flat through $k{=}32$
(${\leq}2$\,pp); (ii)~$S{=}1$ rises $0.20{\to}0.34$ then plateaus---treat $k$
as a \emph{privacy knob} on long multi-session histories after key-fact
summarization; (iii)~$S{=}0$ saturates near $0.93$ by $k{\geq}12$, not
$1.0$.
\textbf{Application:} use Fig.~\ref{fig:locomo_wide_k} to stress-test
post-summarization privacy under wide retrieval; use LME $N{=}500$ for
utility-calibrated operating points.

\section{Statistical Analysis}
\label{app:stats}

All CIs use percentile bootstrap over per-instance booleans
($n{=}1000$ resamples). Cross-$S$ comparisons are
\emph{paired}: each resample reuses the same drawn instance indices for
both $S$ arms, so per-instance correlation narrows the CI on the
difference relative to resampling each arm separately. On Gemma at
$N{=}500$, $k{=}3$,
$S{=}0{\to}1$ yields the bulk of privacy gain: AER $0.83{\to}0.30$
(${-}53$\,pp) vs.\ cosine PR $0.60{\to}0.55$ (${-}5$\,pp) and PR\_fact
$0.46{\to}0.45$ (${-}1$\,pp).
The $S{=}0$ slice at $k{=}3$ ($0.83$) exceeds the $k$-averaged $0.73$ in
Table~\ref{tab:main} because AER rises with retrieval breadth under
raw memory ($0.47/0.83/0.90$ at $k{=}1/3/6$).
Gain $S{=}1{\to}2$ is marginal for headline AER
(pooled ${\approx}2$\,pp; CI overlaps zero, $p{\approx}0.07$ at $k{=}3$) but
significant for PR\_fact ($p{<}0.001$), driven mainly by token-fact loss
(Fig.~\ref{fig:pr_fact_breakdown}).


\section{Utility Measurement}
\label{app:mechanistic}

Personalization recall is scored at three strictness levels (not three
estimators of one quantity):

\noindent
{\footnotesize
\setlength{\tabcolsep}{3pt}
\renewcommand{\arraystretch}{1.05}
\begin{tabular}{@{}p{0.24\columnwidth}p{0.52\columnwidth}p{0.18\columnwidth}@{}}
\toprule
\textbf{Metric} & \textbf{Criterion} & \textbf{Level} \\
\midrule
Cosine PR & Emb.\ sim.\ ${>}0.50$ or keyword hit & Lenient \\
LLM-judge & Holistic YES/NO on answer & Mid. \\
PR\_fact & Mean over $2$--$3$ atomic facts & Strict \\
\bottomrule
\end{tabular}
}

\begin{figure}[t]
  \centering
  \IfFileExists{figures/n500/pr_fact_breakdown.pdf}{%
    \includegraphics[width=0.78\columnwidth]{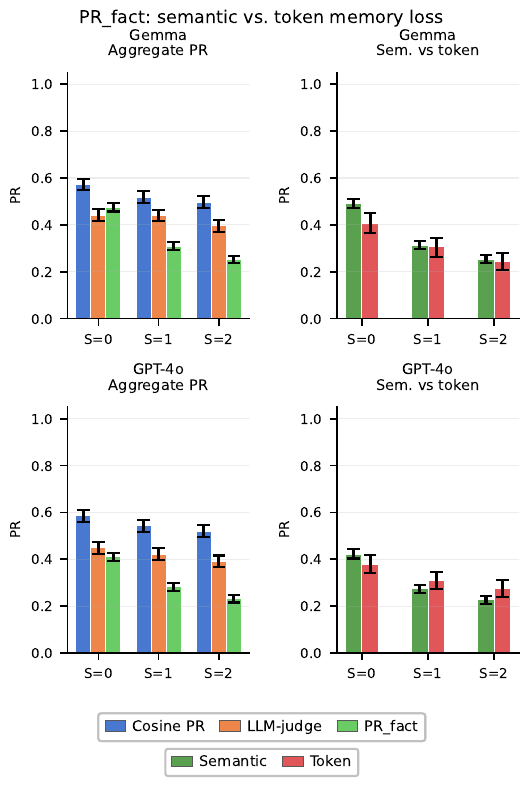}%
  }{%
    \fbox{\parbox{\columnwidth}{\centering\vspace{1cm}%
      [figure pending]\vspace{1cm}}}%
  }
  \caption{\textbf{PR\_fact breakdown} (Gemma/GPT-4o-mini, $N{=}500$, pooled
    over $k$; unified \texttt{gpt-4.1-mini} judge).
    Two rows per model; columns $=$ $S\in\{0,1,2\}$.
    \textbf{Left (aggregate):} cosine PR drops ${\approx}4$--$7$\,pp
    $S{=}0{\to}S{\geq}1$; LLM-judge ${\approx}4$--$6$\,pp; PR\_fact
    ${\approx}0$--$2$\,pp (Gemma) / slightly \emph{up} on GPT-4o-mini
    (${\approx}0.39{\to}0.40$ at $S{=}1$).
    \textbf{Right (semantic vs.\ token):} semantic recall is flat across $S$
    (Gemma ${\approx}0.44{\to}0.43$); token recall falls
    (${\approx}0.40{\to}0.37{\to}0.32$ Gemma;
    ${\approx}0.37{\to}0.34{\to}0.32$ GPT-4o-mini).
    Token tags apply on ${\approx}32\%$ of judged rows---laundering removes
    high-entropy verbatim strings while paraphrasable facts persist.}
  \label{fig:pr_fact_breakdown}
\end{figure}

On Gemma at $S{=}1$, $k{=}3$: cosine${=}0.55$, judge${=}0.43$, PR\_fact${=}0.45$.
\textbf{19\%} of rows pass the holistic judge (${\geq}0.5$) but fail
PR\_fact${<}0.5$ --- holistic scores can mask dropped facts after
summarization. Privacy--utility leverage under PR\_fact remains strongly
asymmetric at $k{=}3$ (53\,pp AER gain vs.\ ${\approx}1$\,pp PR\_fact loss);
token-tagged facts (${\approx}32\%$ of judged rows) account for most of the
$S{=}0{\to}1$ PR\_fact gap while semantic recall stays flat
(Fig.~\ref{fig:pr_fact_breakdown}).

\section{Judge Protocol}
\label{app:judge}

\paragraph{Independent judge.}
Utility scores use a judge \emph{outside} the model under test: both Gemma
and GPT-4o-mini answers are judged by \texttt{gpt-4.1-mini} for utility and
PR\_fact. Each judged row records \texttt{judge\_model\_used} /
\texttt{fact\_judge\_model\_used}.

\paragraph{Agreement.}
Cohen's $\kappa$ between cosine and LLM-judge PR is $0.54$ (Gemma) and
$0.50$ (GPT-4o-mini) over 4{,}500 judged rows --- moderate agreement
reflecting different scoring questions (embedding similarity vs.\ holistic
conveyance), not pipeline inconsistency. An 18-case human spot-check on the
cosine path yields $\kappa{\approx}0.44$; disagreements cluster on
knowledge-update and preference types where PR\_fact and cosine diverge most.

\paragraph{Prompts.}
Holistic judge (temperature${=}0$) asks whether RESPONSE conveys the same
factual content as EXPECTED (YES/NO). PR\_fact uses one batched call per
row over $2$--$3$ facts tagged \texttt{SEMANTIC} (paraphrase-OK) or
\texttt{TOKEN} (verbatim) at extraction time.

\end{document}